\documentclass[lettersize,journal]{IEEEtran}
\usepackage{amsmath,amsfonts}
\usepackage{algorithmic}
\usepackage{algorithm}
\usepackage{array}
\usepackage[caption=false,font=normalsize,labelfont=sf,textfont=sf]{subfig}
\usepackage{textcomp}
\usepackage{stfloats}
\usepackage{url}
\usepackage{verbatim}
\usepackage{graphicx}
\usepackage{cite}
\begin{document}
\bstctlcite{IEEEexample:BSTcontrol}

\title{Estimating the treatment effect over time under general interference through deep learner integrated TMLE}

\author{Suhan Guo, Furao Shen, Ni Li
\thanks{S. Guo is with the National Key Laboratory for Novel Software Technology, Nanjing University, China, and also with the School
of Artificial Intelligence, Nanjing University, Nanjing, China.
Email: shguo@smail.nju.edu.cn}
\thanks{F. Shen is with the National Key Laboratory for Novel Software Technology, Nanjing University, China, and also with the School
of Artificial Intelligence, Nanjing University, Nanjing, China.
Email: frshen@nju.edu.cn}
\thanks{Office of Cancer Screening, National Cancer Center/ National Clinical Research Center for Cancer/Cancer Hospital, Chinese Academy of Medical Sciences and Peking Union Medical College, Beijing, China.
Email: nli@cicams.ac.cn}}




\maketitle

\begin{abstract}
Understanding the effects of quarantine policies in populations with underlying social networks is crucial for public health, yet most causal inference methods fail here due to their assumption of independent individuals. We introduce DeepNetTMLE, a deep-learning-enhanced Targeted Maximum Likelihood Estimation (TMLE) method designed to estimate time-sensitive treatment effects in observational data. DeepNetTMLE mitigates bias from time-varying confounders under general interference by incorporating a temporal module and domain adversarial training to build intervention-invariant representations. This process removes associations between current treatments and historical variables, while the targeting step maintains the bias-variance trade-off, enhancing the reliability of counterfactual predictions. Using simulations of a ``Susceptible-Infected-Recovered'' model with varied quarantine coverages, we show that DeepNetTMLE achieves lower bias and more precise confidence intervals in counterfactual estimates, enabling optimal quarantine recommendations within budget constraints, surpassing state-of-the-art methods.
\end{abstract}


\section{Introduction}
\IEEEPARstart{T}{o} mitigate the spread of infectious diseases, public health decision-makers face a challenging health-economy trade-off \cite{pangalloUnequalEffectsHealth2024}: balancing strict shelter-in-place policies to prevent outbreaks with the economic costs and societal impacts of such measures. Accurate estimation of the effects of varying quarantine coverages is therefore crucial, as it enables policymakers to design effective strategies that minimize both health risks and economic disruption. While randomized controlled trials (RCTs) are the gold standard for causal inference, they are often costly, time-consuming, and, in some cases, impractical, especially for interventions applied at the network connectivity \cite{boothRandomisedControlledTrials2014}. Leveraging the wealth of social network data and related health information from observational studies offers a feasible and scalable alternative for estimating these treatment effects in real-world conditions.

A substantial body of work \cite{hartfordDeepIVFlexible2017, wagerEstimationInferenceHeterogeneous2018, yoonGANITEEstimationIndividualized2018, guyonBayesianInferenceIndividualized2017, shiAdaptingNeuralNetworks2019} has been developed under the Stable Unit Treatment Value Assumption (SUTVA) \cite{rubinRandomizationAnalysisExperimental1980}, which assumes that an individual’s potential outcome is unaffected by the treatment status of their neighbors. However, this assumption breaks down in social network settings, where the interactions between connected individuals are central to the intervention itself. Some methods have been introduced to address network interference, though they primarily focus on static settings \cite{maCausalInferenceNetworked2021, cortezStaggeredRolloutDesigns2022, zivichTargetedMaximumLikelihood2022}. However, single-time-point interventions may not always reflect real-world conditions, as interventions often evolve alongside the progression of an epidemic. For instance, a shelter-in-place order might be lifted after the disease’s incubation period if no new cases are detected. These time-varying interventions introduce time-varying confounding, where patients’ histories—including their covariates and responses to past interventions—inform future intervention decisions. Addressing these challenges requires models that can handle both network interference and time-varying interventions to assess intervention effectiveness in real-world, dynamic scenarios accurately \cite{mansourniaHandlingTimeVarying2017}. Estimating treatment effects across varying network connectivities over time presents valuable opportunities, such as optimizing quarantine effectiveness, assessing how network size influences treatment impact, and identifying optimal coverage levels and timing for quarantine orders. These insights can enable decision-makers to craft more effective, data-driven policies that optimize health outcomes and resource allocation.

\begin{figure}[!t]
\centering
\includegraphics[width=\linewidth]{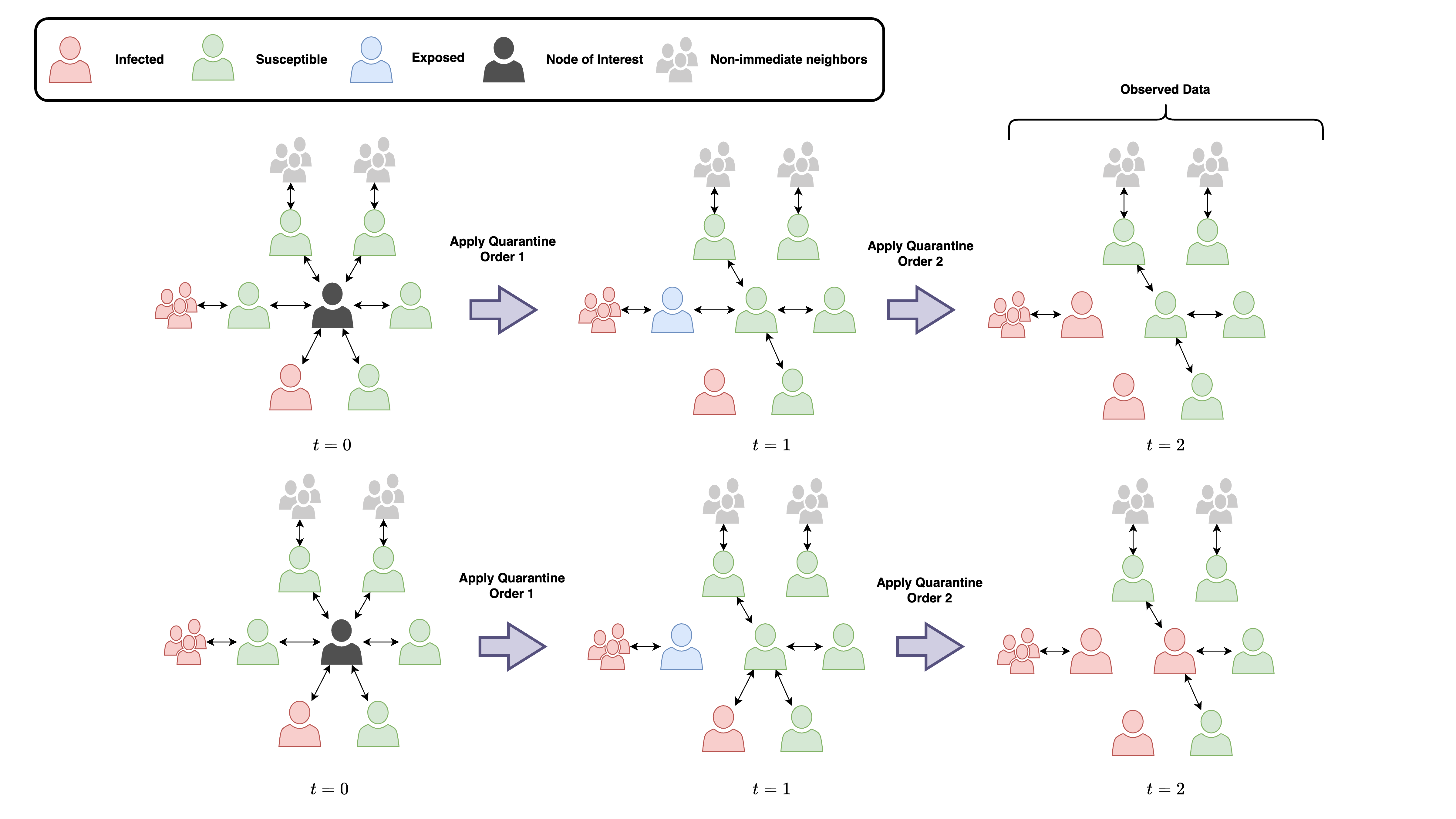}
\caption{Current challenge for estimating causal effect under interference with time-dependent intervention assignments. By changing the order of quarantine treatment, the potential outcome differs. The cross-sectional analysis at the last time point fails to illustrate the difference between two treatment plans.}
\label{fig:current_challenge}
\end{figure}

The biggest challenge when estimating the effects of intervention on the network involves correctly handling the time-dependent network intervention through the change in treatment \cite{vanderlaanTargetedLearningData2018}. 
As illustrated in Figure \ref{fig:current_challenge}, if a model only accounts for the last time step, it may miss critical historical connections and mistakenly attribute observed outcomes to current network structures rather than to previous exposures. For instance, consider individual A, a node with no immediate infected neighbors in the current time step. However, data from earlier time steps reveal that individual A was previously connected to multiple infected nodes and became infected before these connections were removed through quarantine. Lacking this historical context, we might incorrectly infer that the intervention in network structure had a harmful effect.

Existing methods for causal inference on network structure interventions, typically designed for static settings, are inadequate for such longitudinal contexts. These methods rely on single-time-point intervention setups and observational data limited to the last time step, making them ill-suited for estimating counterfactual outcomes where epidemic progression and temporal dependencies are key factors. To effectively capture epidemic dynamics and enable accurate counterfactual prediction, causal inference methods designed for network interference must be adapted to account for longitudinal data.

Time-varying interventions are inherently present in observational data, as public health officials implement mitigation strategies that adapt to the evolving threat of public health hazards \cite{corderFactorsMotivatingTiming2020}. However, these policies embedded in the observational data can introduce significant bias to deep learning models, diminishing their ability to estimate counterfactuals under alternative intervention policies accurately. Addressing these challenges requires developing robust models that can account for both the temporal dependencies and networked nature of interventions, ultimately improving the reliability of intervention-effect estimates and informing public health decision-making.

To address time-varying confounding, traditional methods such as Marginal Structural Models (MSMs) \cite{robinsMarginalStructuralModels2000, mansourniaEffectPhysicalActivity2012} use the inverse probability of treatment weighting (IPTW), creating a pseudo-population that removes dependencies between treatment probabilities and time-varying confounders. However, MSMs can yield poor estimates if IPTW contains extreme weights or the model is misspecified. Machine learning approaches like R-MSN \cite{limForecastingTreatmentResponses2018}, CRN \cite{bicaEstimatingCounterfactualTreatment2020}, and Causal Transformer \cite{melnychukCausalTransformerEstimating2022} apply recurrent neural networks or transformer models to manage temporal correlations, but they assume no interference among individuals and focus on sequence prediction of treatment effects, which is not our primary focus. Targeted Maximum Likelihood Estimation (TMLE), adapted for social network data, introduces a targeting step that optimizes the bias-variance tradeoff for the parameter, ensuring doubly robust estimation for integrating machine learning techniques \cite{schulerTargetedMaximumLikelihood2017}. There have also been attempts to extend the TMLE method with transformer models to handle the longitudinal setting, but their major focus is to facilitate statistical inference in longitudinal settings concerning survival outcomes under dynamic interventions \cite{shirakawaLongitudinalTargetedMinimum2024}.

We introduce DeepNetTMLE, a novel deep learning-based Targeted Maximum Likelihood Estimation (TMLE) approach designed to estimate treatment effects with time-dependent covariates under conditions of general interference. Our main contributions are as follows:

\textbf{Counterfactual Estimation of Social Network Connectivity Interventions.} To estimate counterfactual outcomes for Non-Pharmaceutical Interventions (NPIs), such as quarantine orders issued during pandemics to block human-to-human transmission, we integrate deep learning networks with the well-established TMLE method, preserving its statistical rigor in determining causal effects from observational data. TMLE typically involves four steps: estimating the outcome model, estimating weights, targeting, and estimating the causal effect. DeepNetTMLE innovates by replacing the outcome model with a deep learner and refining the targeting step to enhance prediction accuracy. Through counterfactual estimation of intervention outcomes, DeepNetTMLE can address critical public health questions, such as determining the most cost-effective quarantine coverage and optimizing quarantine budgets, without tracking the initially infected individuals.

\textbf{Domain-Invariant Representations Over Time.} DeepNetTMLE constructs intervention-invariant representations throughout the progression of infectious diseases, effectively breaking the association between historical network connectivity and intervention assignments to eliminate bias from time-dependent confounders. Using domain adversarial training \cite{ganinDomainAdversarialTrainingNeural2016, liDeepDomainGeneralization2018, sebagMultidomainAdversarialLearning2019}, DeepNetTMLE trains the TMLE outcome model to create domain-invariant representations for counterfactual estimation. Experimental results demonstrate that these representations successfully mitigate bias from time-varying confounders, enabling reliable counterfactual outcome estimation. This work is the first to apply domain adaptation for treatment effect estimation under general interference, presenting a novel approach to handling time-varying confounders in this context.

\textbf{Disease-Agnostic Counterfactual Data Sampling.} Estimating counterfactuals typically requires a counterfactual dataset, which is often unattainable in real-world scenarios. For DeepNetTMLE, we developed a data generation mechanism that relies solely on the probability of exposure as input, allowing the network to infer disease-related information directly from observational data. This capability enables DeepNetTMLE to be applied to emerging infectious diseases using only observational data, without requiring detailed disease-specific information that may not be readily available.

In our experiments, we evaluate DeepNetTMLE in a realistic setup using a Susceptible-Infected-Recovered (SIR) model to simulate the epidemic dynamics modified from \cite{zivichTargetedMaximumLikelihood2022}. We demonstrate that DeepNetTMLE outperforms current state-of-the-art methods in predicting counterfactual outcomes, especially as interventions become more stringent. Such causal effects are well-suited for testing in simulation environments, given the high financial costs of implementing these interventions in real-world scenarios.

\section{Related Works}

\subsection{TMLE for Social Network Data}
\noindent Two primary techniques for estimating the average treatment effect (ATE) in causal inference are G-computation \cite{robinsNewApproachCausal1986} and Inverse Probability Weighting (IPW) \cite{robinsMarginalStructuralModels2000}. However, both G-computation and IPW estimators require the correct model specifications to yield consistent estimates and lack theoretical guarantees for normal distribution, making variance estimation for confidence intervals challenging when using flexible machine-learning models. In contrast, Targeted Maximum Likelihood Estimation (TMLE) leverages asymptotically linear estimation and the efficient influence function (EIF) to construct an estimator with optimal statistical properties. This approach ensures that the final estimator is consistent if either the outcome model or the intervention probability model is correctly specified, achieving double robustness \cite{schulerTargetedMaximumLikelihood2017}.

When the potential outcome of one individual is influenced by another’s treatment assignment, the standard no-interference assumption in most causal inference methods is violated. A most prominent example would be the vaccination for infectious diseases, which creates an indirect effect of treatment, defined as the treatment's effect on connected individuals, in addition to the direct effect, defined as the effect of the treatment on his/her own outcome \cite{halloranStudyDesignsDependent1991}. Using contagion as the motivating example, works have been developed to establish the theoretical ground of assessing causal effects for social network data under weak dependence assumption \cite{ogburnCausalInferenceSocial2024, ogburnVaccinesContagionSocial2017}. TMLE has been proposed to estimate the doubly robust causal inference under general interference in observational studies \cite{roseIntroductionTMLE2011, laanTargetedMaximumLikelihood2006}. Maximum likelihood estimation seeks to minimize a global measure (e.g., mean squared error), while the targeting step is introduced to achieve an optimal bias-variance trade-off. 
TMLE has been extended to various contexts, including randomized trials \cite{balzerAdaptivePairmatchingRandomized2015}, stochastic interventions for a single time-point observational data \cite{diazAssessingCausalEffect2013}, complex observational longitudinal studies \cite{neugebauerTargetedLearningRealworld2014}, and social network observational data \cite{zivichTargetedMaximumLikelihood2022}.

\subsection{Treatment Effect Over Time}
\noindent The need to estimate the average effects of time-varying treatments originated in epidemiology, with early methods like G-computation, marginal structural models (MSMs), and structural nested models addressing this demand \cite{robinsNewApproachCausal1986, robinsMarginalStructuralModels2000, hernanMarginalStructuralModels2000}. Since linear regression approaches struggle with complex temporal dependencies, Bayesian non-parametric methods are proposed \cite{xuNonparametricBayesianApproach2016a, schulamReliableDecisionSupport2017, soleimaniTreatmentResponseModelsCounterfactual2017}, which are not compatible with multi-dimensional outcomes and static covariates. 

In the potential outcomes framework, methods such as recurrent marginal structural networks (RMSNs) \cite{limForecastingTreatmentResponses2018}, counterfactual recurrent network (CRN) \cite{bicaEstimatingCounterfactualTreatment2020}, G-Net \cite{liGnetRecurrentNetwork2021}, and Causal Transformer \cite{melnychukCausalTransformerEstimating2022} are introduced to handle biases from time-varying confounders. RMSNs enhance MSMs with recurrent neural networks to estimate the inverse probability of treatment weighting (IPTW) for forecasting treatment responses. CRN reduces the link between patient covariate history and treatment assignment by generating balanced representations that only predict outcomes. G-Net applies g-computation with LSTMs to predict counterfactual outcomes of time-varying treatments under alternative dynamic strategies. Causal Transformer uses the transformer architecture to model time dependencies, employing a counterfactual domain confusion loss to correct for biases from time-varying confounders. There have also been efforts to extend TMLE with transformer models to address longitudinal settings. However, the primary focus of these works has been to support statistical inference for survival outcomes under dynamic interventions, without considering interference \cite{shirakawaLongitudinalTargetedMinimum2024}.

\subsection{Intervention-Invariant Representations for Treatment Effect Estimation}
\noindent Considerable research has been dedicated to obtaining distribution-invariant representations for control and treatment groups to estimate counterfactual outcomes \cite{alaaBayesianNonparametricCausal2018, wagerEstimationInferenceHeterogeneous2018, curthNonparametricEstimationHeterogeneous2021, shalitEstimatingIndividualTreatment2017, liMatchingBalancedNonlinear2017, yaoRepresentationLearningTreatment2018}. Most of them focus on static settings, and some have adapted deep learning algorithms to the estimation \cite{johanssonLearningRepresentationsCounterfactual2016, yoonGANITEEstimationIndividualized2018}. Works including CRN and Causal Transformer, create temporally balanced representations to mitigate bias and estimate counterfactual outcomes in sequential treatment scenarios. However, to develop intervention-invariant representations under general interference and time-varying confounding, sequential generation of treatment and outcome data poses challenges. This is because treatment updates may not occur at every step, and outcome data is typically available only at the final time step.

Domain adaption is the process of transferring a model’s knowledge to a different yet related data distribution. Typically, a model is trained on a source dataset and applied to an unlabeled target dataset. Domain alignment is generally achieved through one of three primary approaches: discrepancy-based, adversarial-based, and reconstruction-based methods \cite{hassanpourzonooziSurveyAdversarialDomain2023}. Discrepancy-based methods seek to minimize divergence by measuring divergence or distance between the distributions. Reconstruction-based methods are first proposed by Ghifary et al. \cite{ghifaryDeepReconstructionclassificationNetworks2016} that a representation is learned to accurately classify the labeled source data and reconstruct source and target data. Adversarial-based methods use adversarial techniques to align distributions by extracting features that effectively classify labeled source data but appear indistinguishable between source and target domains. This category includes state-of-the-art approaches like DANN \cite{ganinDomainAdversarialTrainingNeural2016} and ADDA \cite{tzengAdversarialDiscriminativeDomain2017}.  DANN employs a gradient reversal layer to extract domain-invariant features, while ADDA uses adversarial adaptation with a GAN loss to align a pre-trained source classifier with the target distribution. To prevent the negative transfer, multi-adversarial domain adaptation (MADA) extended the domain discriminator to each class to reduce the negative transform and prevent the wrong classification \cite{peiMultiadversarialDomainAdaptation2018}. Dual Adversarial Domain Adaptation (DADA) simultaneously performs domain-level and class-level alignment using a common discriminator \cite{duDualAdversarialDomain2020}. 

\section{Problem Definition}

\noindent An adjacency matrix $\mathbb{G} = (V, E)$  is used to represent the social network of the population of interest, where $|V| = N$ represents the $N$ individuals and $|E| = M$ denotes $M$ edges between individual pairs. Unlike the typical adjacency matrix, a self-loop is excluded from $\mathbb{G}$ because interference is considered only between different individuals. $\mathbb{G}_{ij}=1$ indicates that there is a connection between individual $i$ and $j$, whereas $\mathbb{G}_{ij}=0$ denotes the absence of connection. Given that the network structure is dynamic, we define the network over time as $\mathbb{G}(t) = [\gamma(0), \dots, \gamma(T)]$.

We denote the observed outcome for individual $i$ over $T$ time steps as $[\upsilon_i(0), \dots, \upsilon_i(T)]$. To account for interference, both an individual’s own exposure and the exposure of others in the population must be considered. The exposure for an individual is defined as $A_i(t) = [\alpha_i(0), \dots \alpha_i(T)]$, and the exposure of others is represented as a summary measure from the immediate contacts according to $\mathbb{G}$, given by:
\begin{align}
    &A_i^s(t) = [\alpha_i^s(0), \dots, \alpha_i^s(T)] \nonumber \\
    &= \left[\sum_{j=1}^{N} \mathbb{I}(\alpha_j(0) = 1) \gamma_{ij}(0), \dots, \sum_{j=1}^{N} \mathbb{I}(\alpha_j(T) = 1) \gamma_{ij}(T)\right].
\end{align}

For each individual $i$, the baseline covariates are represented by $X_i = [\xi_i(0), \dots, \xi_i(T)]$, capturing both static and dynamic characteristics. The static covariates remain constant across time, while dynamic covariates are updated at regular intervals to reflect any changes over time. We further define $X_i^s = [\xi_i^s(0), \dots, \xi_i^s(T)]$ as the summary covariates corresponding to the baseline covariates. These summary covariates are recalibrated at each time step to ensure they remain aligned with the ground truth quarantine measure. To avoid further complicating the notation, we drop the temporal dimension in the following illustration unless otherwise mentioned.

Let $Y_i(a_i, a_{/i})$ represent the potential outcome for individual $i$, where  $a_i$ denotes the exposure of individual $i$, and $a_{/i}$ refers to the exposure statuses of all other individuals in the population, excluding individual $i$. In this model, we assume that the exposure statuses from an individual’s immediate contacts are sufficient to describe the interference effects on their potential outcome. Therefore, we simplify the potential outcome to $Y_i(a_i, a_i^s)$, where  $a_i \in \mathbb{A} = \{0, 1\}$  denotes the treatment assignment (e.g., whether an individual is exposed to a treatment or intervention), and $a_i^s \in \mathbb{A}^s$ captures the summary of the exposures from the immediate contacts of individual $i$, reflecting the indirect treatment effects. This weak dependence assumption allows us to model the problem in a more tractable way, where the exposure status of an individual’s contacts $a_i^s$ is used to characterize the interference effects. The goal of this study is to estimate the average outcome under a given policy $\omega$ for the population. Specifically, we aim to estimate the following causal quantity:
\begin{align}
    \psi = \frac{1}{N} \sum\limits_{i=1}^{N} \mathbb{E}\left[ Y_i(a_i, a_i^s)Pr^{*}(A_i=a_i, A_i^s=a_i^s | X_i, X_i^s, \mathbb{G})\right],
\end{align}
where $Pr^{*}$ denotes the probability under policy $\omega$, and this probability must be estimated based on available data, as it is not directly observed. The expression reflects the expectation of the potential outcome $Y_i(a_i, a_i^s)$ under the intervention (or treatment) $a_i$, weighted by the probability of the exposure assignment $a_i$ and the associated exposures $a_i^s$, conditional on the individual covariates $X_i$, the summary covariates $X_i^s$, and the network structure $\mathbb{G}$.

To ensure the causality in the estimation of $\psi$, three assumptions are made, including causal consistency, conditional exchangeability, and positivity. Details can be found in Supplementary Section I.

In a real-world setting, when estimating the average outcome $\psi$ under policy $\omega$, we lack access to the counterfactual ground truth. This absence prevents us from using counterfactual information to directly fit the model, as it remains unobservable outside of simulations. However, in our simulation experiments, we can observe the counterfactual ground truth. To realistically reflect practical application constraints, we exclude this ground truth data during the training phase of the model and instead, reserve it solely for evaluating the performance of our proposed algorithm.

To assess the performance, we define the following metrics:
\textbf{Bias:} The mean difference between the estimated outcome and the ground truth across $U$ simulation processes, calculated as:
    \begin{align}
        \text{Bias} = \frac{1}{U}\sum\limits_{k=1}^{U} (\hat{\psi}_k - \psi_k).
    \end{align}
\textbf{Empirical Standard Error (ESE):} The standard deviation of the simulation estimates for each policy. Let $b_k = \hat{\psi}_k - \psi_k$, ESE is defined as:
    \begin{align}
        \text{ESE} = \sqrt{\frac{\sum_{k=1}^{U} (b_k - \text{Bias})^2}{U}}.
    \end{align}
\textbf{95\% Confidence Interval (CI) Coverage:} The proportion of 95\% CIs containing the true mean of the outcome. Defining the lower and upper bound of CI as LCI and UCI, we have,
    \begin{align}
    \text{LCI} = \hat{\psi}_k - z_{1-\frac{\alpha}{2}} * \sqrt{{\frac{\hat{\sigma}_{k}^{2}}{N}}}, \\
    \text{UCI} = \hat{\psi}_k + z_{1-\frac{\alpha}{2}} * \sqrt{{\frac{\hat{\sigma}_{k}^{2}}{N}}}, 
    \end{align}
    where $\hat{\sigma}^{2}$ is the estimated variance, The $z_{1-\frac{\alpha}{2}}$ is the $1-\frac{\alpha}{2}$ quantile of a standard normal distribution, and $\alpha = 0.05$ for 95\% CI. 
    The 95\% CI coverage is then calculated as:
    \begin{align}
    \text{Cover}_k &= 
        \begin{cases}
            1, & \mbox{LCI} < \psi \mbox{~and~} \mbox{UCI} > \psi \\
            0, & \mbox{Otherwise},
        \end{cases} \nonumber \\
    \text{Cover} &= \frac{1}{U} \sum\limits_{k=1}^{U} \text{Cover}_k.
    \end{align}
In this evaluation framework, a smaller Bias and ESE indicate a more accurate and precise estimation of the average outcome, while a higher 95\% CI coverage suggests that the confidence intervals are capturing the true mean of the outcome more consistently, reflecting a reliable estimation process. Details on the variance estimation methodology are provided in Section~\ref{subsec:estimate_variance}.

\section{DeepNetTMLE}
\noindent To estimate $\hat{\psi}$, the DLNetworkTMLE consists of five steps: 
\begin{enumerate}
    \item Estimate the outcome model $f(\cdot)$
    \item Estimate Inverse Probability of Treatment Weighting (IPTW) with exposure model $g(\cdot)$ and summary exposure model $h(\cdot)$
    \item Estimate the correcting intercept $\epsilon$, also known as ``targeting''
    \item Estimate the $\hat{\psi}$ using Monte Carlo approach
    \item Estimate the direct variance $\hat{\sigma}_{d}$ and latent variance $\hat{\sigma}_{l}$
\end{enumerate}
In this approach, models estimated from the observed data are denoted by $f(\cdot), g(\cdot), h(\cdot)$, while those estimated with Monte Carlo-sampled data are denoted with an asterisk (e.g. $f^{*}(\cdot), g^{*}(\cdot), h^{*}(\cdot)$). This five-step estimation process corresponds to a single simulation, and for simplicity, the subscript $k$, denoting a specific simulation iteration, is omitted.

\subsection{Estimate the outcome model}
\noindent The outcome model in this framework is designed to predict the outcome variable based on exposure variables and covariates that reflect temporal dynamics in population exposure. For the deep learning extension of networkTMLE, exposures and covariates across selected time steps are incorporated into the model, enabling it to capture changes in exposure across the population over time. This is expressed as:
\begin{align}
    &\mathbb{E}\left[ Y_i(t) | A_i(t), A_i^s(t), X_i, X_i^s, \mathbb{G} \right] \nonumber\\
    &= f(A_i(t), A_i^s(t), X_i, X_i^s, \mathbb{G}), 
\end{align}
where the function $f(\cdot)$ represents the model that estimates the expected outcome $Y_i(t)$ given individual and summary exposures, covariates, and network structure.

The first step involves estimating this outcome model using observed data, which we consider as a representative sample from the broader network super-population. Traditional models, such as the Generalized Linear Model (GLM), struggle to handle temporal dependencies. Therefore, in the absence of a deep learning model, only data from the final time step is used to estimate the outcome $\hat{Y}_i(T)$. To ensure compatibility with the GLM approach, we retain only the final estimate, denoted as $\hat{Y}_i(T) = c(f(\cdot))$. The model obtained here will then serve as a key contributor in Step 4 of the estimation procedure.

\subsection{Estimate IPTW}
\noindent The Inverse Probability of Treatment Weighting (IPTW) weights are introduced to account for dependencies among individuals within the networked data, as outlined in semi-parametric estimation and targeted maximum likelihood methods. The weights are structured to balance observed and sampled data through a decomposition into a numerator and denominator, which are fitted on sampled data and observed data, respectively. This is shown as:
\begin{align}
    W_i = \frac{c(g^{*}(X_i(t), X_i^s(t)))h^{*}(\xi_i(T), \xi_i^s(T))}{c(g(X_i(t), X_i^s(t)))h(\xi_i(T), \xi_i^s(T))}.
\end{align}
In the exposure model, we could potentially utilize a deep learning framework to capture temporal dependencies within the data, and, accordingly, we incorporate a reservation function $c(\cdot)$ at the beginning of the model to account for this. However, in the current study, we have opted to use a Generalized Linear Model (GLM) with a Gaussian distribution as the exposure model. This choice allows us to focus on examining the effects contributed specifically by the deep learning model in the outcome estimation, isolating its impact without additional complexity from a deep learning-based exposure model.

The specific form of the summary exposure $A_i^s$ is determined by the choice of summary measure $S$. For instance, if $S$ represents summation, $A_i^s$ becomes a count variable indicating the exposure among immediate contacts. This structure introduces a class imbalance in the sampled data due to the range of possible values $A_i^s$ can take, which grows incrementally. Consequently, we employ a Generalized Linear Model (GLM) with a Poisson distribution to estimate $A_i^s(T)$ rather than using deep learning models, which may be less effective given the class imbalance.
In this framework, the sampled data refers to $M$ copies generated to represent random draws from a broader super-population, indexed by $l=1, \dots, M$. Starting from the initial time step $t = 0$, the generation process begins by sampling the exposure variable $A_{il}^{*}(1)$ based on a policy-specified probability  $Pr_\omega$ from distribution $\mbox{Bern}(Pr^{*}(A_i(1)=1 | X_i(0), X_i^s(0)) = p_\omega)$. The sampled summary exposure measure for each time step is then recalculated as $A_{il}^{s*}(1) = \mathbb{I}(A_{jl}^{*} = 1) \mathbb{G}_{ij}^{*}(0)$ reflecting the exposure from immediate contacts. For each copy, this sampled data is generated sequentially across time steps, resulting in a final dataset that matches the observed data regarding the number of time steps. Beyond generating exposure variables, other data elements within each simulated policy are created following the same approach outlined in Supplementary Section II. Importantly, the true outcome and exposure models, as well as the identities of initially exposed individuals, are hidden in the process to prevent information leakage that could bias the model evaluation.

\subsection{Targeting}
\noindent The targeting step aims to estimate the offset $\epsilon_i$, which measures the discrepancy between the estimated outcome and the true outcome, acting as a correcting intercept. This offset is obtained by fitting a logistic regression model using weighted maximum likelihood with the estimated Inverse Probability of Treatment Weighting (IPTW) weight $W_i$, shown as:
\begin{align}
    \text{logit}(Pr(Y_i=1)) = \epsilon_i + \text{logit}(\hat{Y}_i).
\end{align}
When deep learning models are used for outcome estimation, predicted outcomes $\hat{Y}_i$ are often dichotomized to $0$s and $1$s by the $\text{sigmoid}(\cdot)$. This transformation can lead to extreme values of $\epsilon_i$, which are implausible. Therefore, for deep learning models, $\hat{Y}_i$ values are clipped within the range $(0.05, 0.95)$, and any resulting out-of-bounds $\epsilon_i$ values are corrected to $0$ with a threshold set to $10$.

\subsection{Estimate the average outcome}
\noindent The average outcome $\hat{\psi}$ should be calculated using the Monte Carlo approach. Hence, the previously sampled data $A_{il}^{*}(t)$ and $A_{il}^{s*}(t)$ for $l = 1, \dots, M$ can be reused to predict the outcome $\hat{Y}_{il}^{*}(t)$ under policy $\omega$. Utilizing the targeting offset $\epsilon_i$ as the correction, we define the procedure as follows:
\begin{align}
    \Tilde{Y}_{il}^{*}(T) &= c(f(A_{il}^{*}(t), A_{il}^{s*}(t), X_i^*, X_i^{s*}) + \epsilon_i), \\
    \hat{\psi} &= \frac{1}{MN} \sum\limits_{l=1}^{M} \sum\limits_{i=1}^{N} \Tilde{Y}_{il}^{*}(T).
\end{align}
Here, $\Tilde{Y}_{il}^{*}(T)$ represents the corrected predicted outcome.

\begin{figure*}
    \centering
    \includegraphics[width=\textwidth]{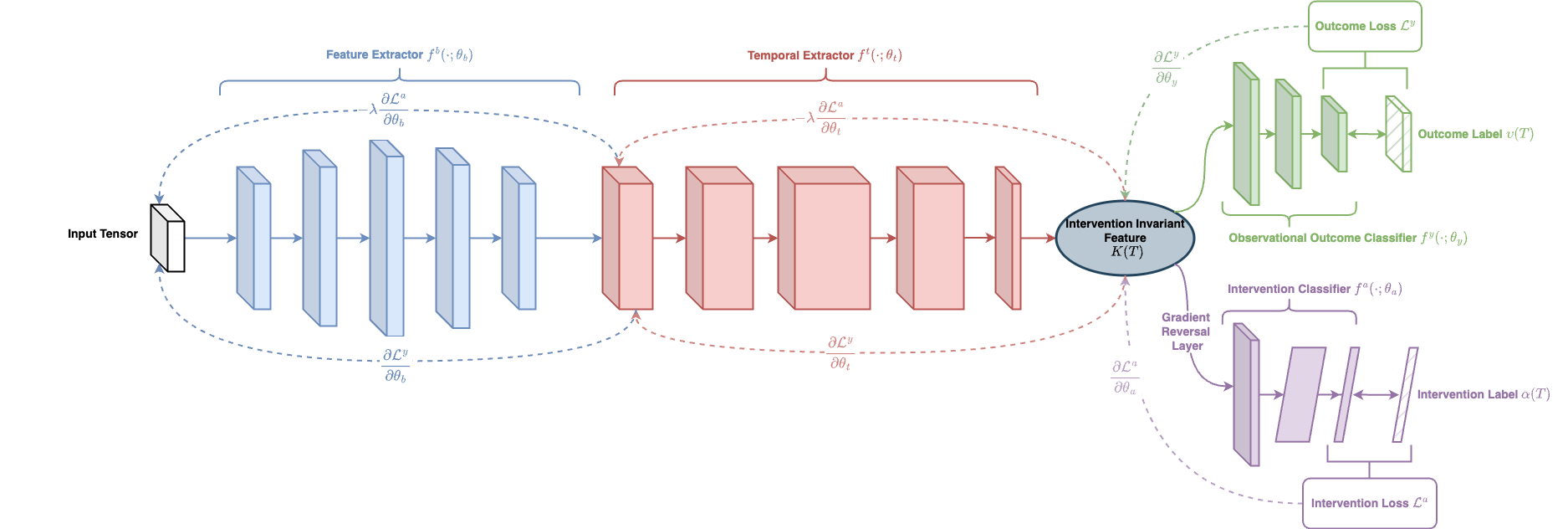}
    \caption{Network structure of outcome model. The solid and dashed lines indicate forward and backward propagation, respectively.}
    \label{fig:network}
\end{figure*}

The network modules in our proposed methods are composed of linear layers, forming the foundational architecture. By incorporating Multi-Layer Perceptrons (MLPs) into different modules within the network, we aim to empirically demonstrate that deep learning models are effective outcome estimators. Additionally, this setup highlights the flexibility of our framework, allowing more specialized network architectures, such as graph neural networks (GNNs), to be integrated seamlessly. This adaptability suggests that our framework can leverage the strengths of various neural network structures depending on the context and data requirements. The network structure is illustrated in Figure \ref{fig:network}.

Dynamic treatment strategies across an observational period introduce time-varying confounding, which can bias outcome estimation. To address this, we construct a treatment-invariant representation for each individual, denoted as $K_i(t) = [\kappa_i(0), \dots, \kappa_i(T)]$, where $Pr(\kappa_i(t) | \alpha_i(t) = 0) = Pr(\kappa_i(t) | \alpha_i(t) = 1)$, for each $t \in [0, T]$. Given access to potential outcomes at each time step, an optimal approach generates $\kappa_i(t=Q)$ for each specific time $Q$, necessitating a sequential design for the network. A significant advantage of sequential networks, particularly those with an encoder-decoder structure, is their capacity to generate multi-step-ahead outcome predictions based on treatment plans. This approach, as used in prior studies, enables the network to anticipate outcomes across multiple time steps, enhancing the robustness of the outcome estimation given dynamic treatments \cite{bicaEstimatingCounterfactualTreatment2020, melnychukCausalTransformerEstimating2022}. For our study, the outcome inference is a point estimate, and the primary outcome of interest is the cumulative infliction of infections throughout the endemic period $T$. This cumulative outcome is not available at intermediate time steps $t \in [0, T-1]$. Hence, the sequential networks are less optimal due to their slow inference speed. To overcome this limitation, instead of treating the different treatments at each time step as distinct domains, we focus on constructing intervention-invariant representations at the final time step by ensuring that $Pr(K_i(T) | A_i(T) = 0) = Pr(K_i(T) | A_i(T) = 1)$. Here, $K_i(T)$ and $A_i(T)$ denotes that the feature is estimated with an intervention implemented over the entire endemic span. To generate $K_i(T)$, we introduce a temporal module that consists of linear projection layers organized in a U-net style. This module maps the temporal information from the input feature $S_i$ to the desired output dimension. The transformation is described as follows:
\begin{align}
    K_i(T) = f^t (S_i(T); \theta_t).
\end{align}
If the temporal span of the input data is equal to 1, this temporal module simplifies to the identity function. The backbone of the network is the feature extraction module, composed of linear layers designed to project features to a target dimension. This module first processes input features and then passes them to the temporal module to aggregate intervention effects along with baseline covariates, ensuring that both the direct and indirect exposures are effectively captured in the final representation. The feature extraction is expressed as:
\begin{align}
    S_i(T) = f^b (X_i(T), X_i^s(T), A_i(T), A_i^s(T); \theta_b).
\end{align}

Presuming that the distribution for different interventions $A_i(T) = 0$ and $A_i(T) = 1$ to be similar but shifted, we would like the downstream structure to be able to distinguish outcomes for the observed intervention while failing to distinguish between samples from observational and counterfactual domains. By organizing the loss function to minimize outcome classification error while maximizing intervention classification error, the model encourages the representations to encode outcome-related features without explicitly encoding intervention specifics. The prediction for outcome $\hat{\upsilon}_i(T)$ and intervention $\hat{\alpha}_i(T)$ are defined as:
\begin{align}
    \hat{\upsilon}_i(T) &= f^y(f^t(f^b (X_i(T), X_i^s(T), A_i(T), A_i^s(T); \theta_b); \theta_t); \theta_y), \\
    \hat{\alpha}_i(T) &= f^a(f^t(f^b (X_i(T), X_i^s(T), A_i(T), A_i^s(T); \theta_b); \theta_t); \theta_a).
\end{align}
This architecture seeks to isolate outcome features while discouraging intervention-related biases. The loss function is composed of two parts, the outcome loss $\mathcal{L}_{i}^{y}$ and the intervention loss $\mathcal{L}_{i}^{a}$, i.e.,
\begin{align}
    \mathcal{L}_{i}^{y}(\theta_b, \theta_t, \theta_y) &= -(\upsilon_i(T)\textit{log}(\hat{\upsilon}_i(T)) \nonumber \\
    &+ (1-\upsilon_i(T))\textit{log}(1-\hat{\upsilon}_i(T))), \\ 
    \mathcal{L}_{i}^{a}(\theta_b, \theta_t, \theta_a) &= -(\alpha_i(T)\textit{log}(\hat{\alpha}_i(T)) \nonumber \\
    &+ (1-\alpha(T))\textit{log}(1-\hat{\alpha}_i(T))), \\
    \mathcal{L}_i &= \sum_{i=1}^{N} \mathcal{L}_{i}^{y}(\theta_b, \theta_t, \theta_y) - \lambda \mathcal{L}_{i}^{a}(\theta_b, \theta_t, \theta_a), \label{eq:loss_fn}
\end{align}
where the hyperparameter $\lambda$ controls the balance between outcome prediction and intervention invariance, increasing exponentially during training to promote this invariance over time \cite{ganinUnsupervisedDomainAdaptation2015}.  By training with Eq. \eqref{eq:loss_fn}, the model seeks the saddle point:
\begin{align}
    (\hat{\theta}_b, \hat{\theta}_t, \hat{\theta}_y) &= \underset{\theta_b, \theta_t, \theta_y}{\text{arg min}} \mathcal{L}_{i}(\theta_b, \theta_t, \theta_y, \hat{\theta}_a) \\
    \hat{\theta}_a &= \underset{\theta_a}{\text{arg max}} \mathcal{L}_{i}(\hat{\theta}_b, \hat{\theta}_t, \hat{\theta}_y, \theta_a).    
\end{align}
To achieve this with backpropagation, we employ a gradient reversal layer (GRL) between the temporal and intervention modules. The GRL behaves as an identity during forward propagation but reverses the gradient sign in backward propagation by multiplying it by $-\lambda$. This allows the model to reach the saddle point using standard backpropagation and stochastic gradient descent (SGD) adapted optimizers, effectively disentangling outcome and intervention representations within the network without having to reinvent optimizers.

\subsection{Estimate the variance} \label{subsec:estimate_variance}
\noindent Two types of variance—direct and latent—are estimated to account for different levels of dependence among observations:
The direct variance is short for the direct transmission variance, which assumes that dependence between observations is fully captured by the measured covariates of immediate contacts, shown as:
\begin{align}
    \hat{\sigma}_{d}^{2} = \frac{1}{N} \sum\limits_{i=1}^{N} \left( W_i*(Y_i - \hat{Y}_{i})\right)^{2}.
\end{align}
On the other hand, the latent variance expands the dependence between observations to the second-order contacts, defined as the immediate neighbors of one's immediate neighbors. The latent variance is estimated by:
\begin{align}
    \hat{\sigma}_{l}^{2} = \frac{1}{N} \sum\limits_{i=1}^{N} \sum\limits_{j=1}^{N} \mathcal{G}_{ij} \left( W_i*(Y_i - \hat{Y}_{i}) * W_j*(Y_j - \hat{Y}_{j})\right),
\end{align}
where $\mathcal{G}_{ij}$ denotes the connectivity between node $i$ and node $j$ and when $i=j$, $\mathcal{G}_{ij}=1$. Incorporating latent variance to account for second-degree network dependencies indeed provides a more realistic representation of infectious disease transmission dynamics, as it captures the influence not only of direct contacts but also of indirect interactions within a network. This approach likely improves the robustness of estimations by reflecting the more complex, layered nature of transmission paths in real-world settings.

\section{Experiments With Quarantine Simulation Model}\label{sec:experiment}
\noindent The experiments are conducted using simulated datasets for infectious diseases. The simulation entails a compartment model, namely the Susceptible-Infected-Recovered (SIR) model, which describes the human-to-human transmission of the infectious agents. The hypothetical exposure measure quarantine compliance represented a ``leaky'' model, such that the quarantine reduces the probability of infection given a single exposure to an infectious agent. The spillover effect of quarantine is composed of the contagion effect, which means quarantined individuals are less likely to become infected and thus less likely to transmit, and infectiousness effects, which means quarantined-but-infected individuals have reduced probability of transmitting the disease. The simulation framework enables access to counterfactual datasets under various policy interventions, allowing us to directly measure the bias between estimates and the counterfactual ground truth.

Six network types were used in the experiments, including uniform random graphs and modified clustered power-law random graphs, each with node counts of $N=500$, $1000$, and $2000$. The uniform random graphs were generated with a uniform degree distribution, ranging from a minimum degree of 1 to a maximum degree of 6. The modified clustered power-law random graphs, on the other hand, were constructed by generating separate subgraphs and then adding random edges between these subgraphs. 

For each simulation, $1\%$ of $N$ individuals in network $\mathbb{G}$ are selected as initially infected individuals. Individuals categorized in the ``Infectious'' compartment are deemed to be actively infectious. Infectious individuals can spread the disease to their immediate contacts based on the assigned transmission probability. Each individual remains infectious for 5 discrete time steps before transitioning to the ``Recovered'' compartment, at which point they are no longer capable of transmitting the disease. Transmission events occur over a period of 10 time steps, with the quarantine period set to $P=2$.

In our study, the exposure (treatment/intervention) corresponds to compliance with quarantine orders, where the quarantined individuals have the edges of their respective nodes removed for the entire quarantine duration. The outcome is defined as the cumulative infliction of disease, representing the total number of infections an individual experiences over the endemic period $T$. We evaluate two types of action plans: full coverage, where the budget allows quarantining 100\% of the population, and partial coverage, where the budget covers 50\% of the population. In the partial coverage scenario, either the most connected or least connected nodes are prioritized for quarantine. For each action plan, we test 19 levels of exposure probability $(p_{\omega})$. Further details on the simulation setup are provided in Supplementary Section II.

\subsection{Benchmark Models}
\noindent To test the robustness of our model, we evaluate performance under four scenarios:
\begin{itemize}
    \item CC (Correct-Correct): Independent variables in both the exposure model and outcome model are correctly specified.
    \item CW (Correct-Wrong): Independent variables are correctly specified in the exposure model but not in the outcome model.
    \item WC (Wrong-Correct): Independent variables are incorrectly specified in the exposure model but correctly specified in the outcome model.
    \item Flexible: In the exposure model, the independent variables are correctly specified, but continuous variables are discretized into categorical variables to address right-skewness. For the outcome models, all available variables are included to allow the model to perform feature selection and identify the most relevant predictors.
\end{itemize}
This study is the first to apply deep learning models to predict causal effects under interference while incorporating temporal effects. For baseline comparisons, we use the Generalized Linear Model (GLM) and the L2-penalized regression model (L2). A binomial link function is used to fit the exposure and outcome models, while a Poisson link function is applied to fit the exposure summary model. The GLM is used in the CC, CW, and WC scenarios, whereas the L2 model is applied in the Flexible scenario, as each level of the categorical variable is treated as an independent predictor.

\subsection{Performance under different policies and scenarios}
\noindent To ensure the robustness and reliability of the results, we ran each experiment 30 times across different network configurations, specifically uniform random graphs with sizes $N = 500$, $1000$, and $2000$, as well as power-law random graphs with sizes $N = 500$ and $1000$. For the power-law random graph with $N = 2000$, we reduced the number of repetitions to 15 due to computational constraints, which is still sufficient to observe meaningful patterns while maintaining efficiency. Each run followed the predefined quarantine strategies for the given network, with particular attention to how quarantine policies impacted disease transmission under varying levels of network connectivity and size. Latent coverage comparisons are available in the Supplementary Section V Figure 3-5.
The accuracy of outcome inference is shown in the Supplementary Section IV-A.

\subsubsection{Strategy: constant with budget covering all individuals}
\begin{figure*}[!h]
\centering
\includegraphics[width=\textwidth]{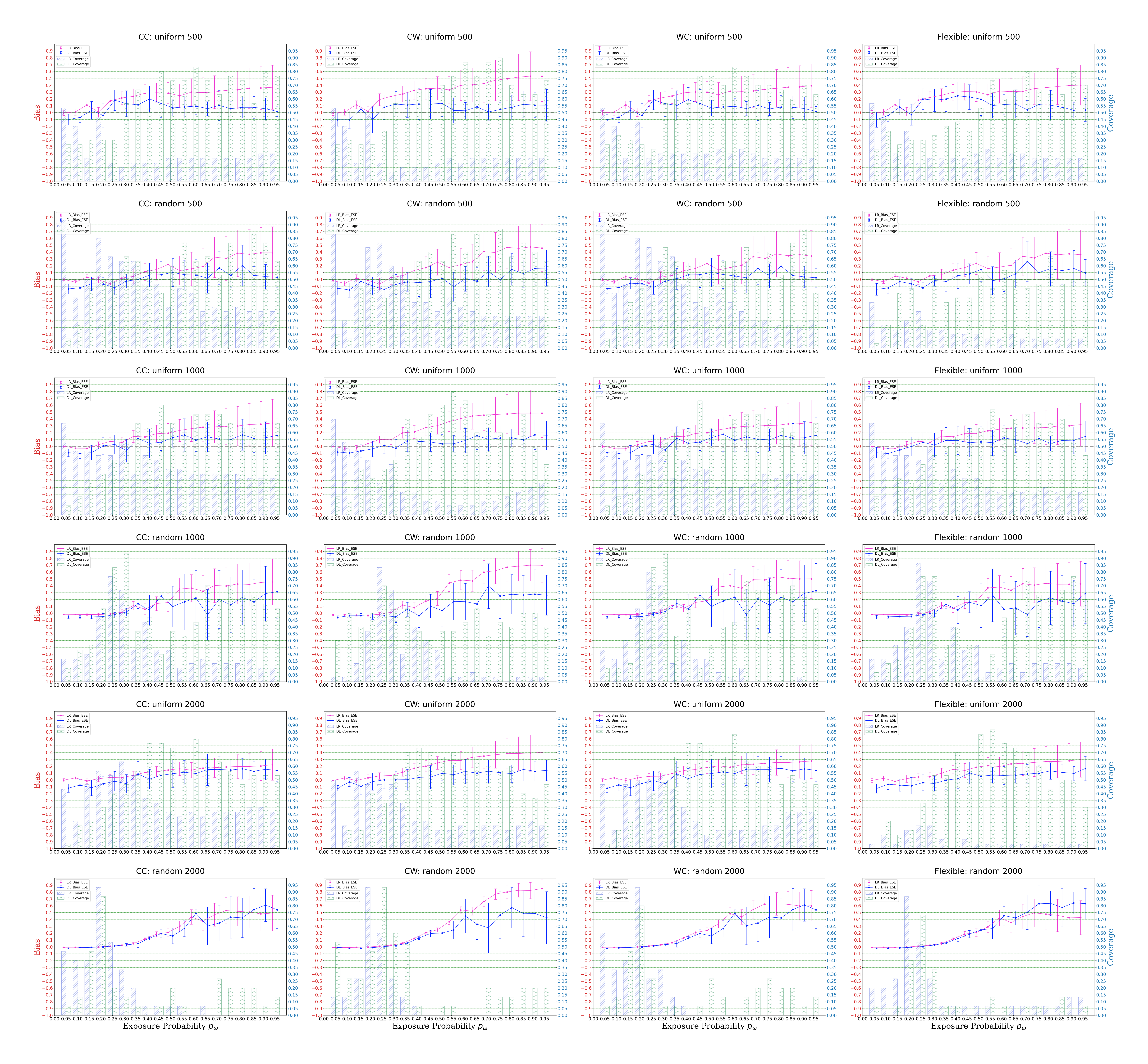}
\caption{Performance comparison between linear regression and deep learning model in predicting the cumulative infliction of disease for uniform and power-law random graph with $N=500, 1000$ and $2000$ under constant strategy with budget covering all population. The deep learning model has a reception field over the last nine time steps of observational data. Bias and ESE are viewed using the left y-axis in red. The cover is shown in blue on the right y-axis. The grey horizontal dash indicated the zero as a benchmark line for bias estimation.}
\label{fig:results_node_all}
\end{figure*}

\noindent The results presented in Figure \ref{fig:results_node_all} showcase the performance of outcome prediction under the constant strategy, where it is assumed that the budget is sufficient to cover all individuals in the network.

In general, we observe that the bias estimated using the deep learner is consistently closer to zero, with a smaller Empirical Standard Error (ESE) compared to those produced by the linear regression models. When the exposure probability is small (i.e., $p_{\omega} < 0.3$), the bias improvement of the deep learner over the linear regression model is marginal. However, as the exposure probability increases (indicating a higher likelihood of quarantine activation), the discrepancy between the sampled counterfactual data and the observed data becomes more pronounced. Consequently, the linear regression model, which is fitted using observed data alone, is more sensitive to this distribution shift. In contrast, the deep learner is trained with both labeled and unlabeled data, leveraging domain adaptation techniques to mitigate the impact of this shift, leading to better performance in cases with higher exposure probabilities. 

An exception is observed under the Flexible scenario for power-law random graphs with $N=2000$, where the benchmark L2 model outperforms deep learners in bias estimation for high exposure probability (i.e., $p_{\omega} > 0.5$). Upon further investigation of the simulated dataset, this phenomenon is linked to the reception field covering an odd number of days. The incubation period in the simulation is set to 2 days, resulting in an adjacency matrix that alternates between loosely connected and densely connected states every 2 days. When the reception field slices the data using an odd number of days, it disrupts the model’s ability to track the most recent incubation dynamics. This loss of crucial temporal context hampers the deep learner’s capacity to accurately predict the counterfactual outcomes, particularly at higher exposure probabilities such as $p_{\omega}=0.95$. This highlights the importance of aligning temporal slicing with the underlying periodicity of the data to ensure effective modeling.

A similar trend is observed in the comparison of coverage. For $p_{\omega} < 0.3$, the coverage from the deep learner does not consistently outperform that of the linear regression. However, as the counterfactual data becomes increasingly distant from the observed data (with higher exposure probabilities), the coverage from the deep learner significantly improves over that of the linear regression model. The latent coverage also follows a comparable trend.

When comparing uniform random graphs to power-law random graphs, we find that, for uniform graphs, counterfactuals can be consistently estimated. As the exposure probability increases, the upper limit of estimated bias for both linear regressions and deep learners remains lower in uniform graphs, with values around 0.5 for linear regressions and 0.2 for deep learners. In contrast, the upper limit of bias for power-law random graphs is noticeably higher (0.9 for linear regression and 0.7 for deep learners). This difference arises because power-law random graphs contain nodes that are more densely connected than uniform graphs. Consequently, higher exposure levels result in greater variations in connectivity, exacerbating the distribution shift between sampled counterfactuals and observed data. Thus, the performance of models trained on power-law random graphs is more volatile, especially at higher exposure probabilities.

The misspecification of independent variables in either the exposure or outcome model did not substantially affect the performance of deep learners. However, linear regression models exhibited a noticeable deterioration in performance under these conditions. Misspecifying the outcome model variables, as demonstrated in Figures \ref{fig:results_node_all}, leads to an increase in the ESE for both deep learning and linear regression models, compared to models fitted with correctly specified exposure and/or outcome model variables.  
The benchmark and our model employ the same Generalized Linear Model (GLM) for the exposure model. However, we observe that misspecification in the exposure model variables can be partially mitigated by the deep learner in the outcome model. Our model consistently outperforms the benchmark, demonstrating improvements in both bias and coverage. This is clearly illustrated in Figures \ref{fig:results_node_all}. Specifically, the deep learner’s ability to adapt to the misalignment in the exposure model enables it to generate more accurate and robust counterfactual predictions than the benchmark GLM. The flexible variables approach performs reasonably well for policies with an exposure probability smaller than 0.5 when using linear regressions. However, the coverage declines significantly for higher exposure probabilities in linear regression models. In the flexible variable scenario, where continuous variables are discretized to handle right-skewness, deep learners are unaffected by this transformation. This is because categorical variables are treated as continuous variables within the model to maintain balance across groups, avoiding sample size imbalances.

\begin{figure*}[!h]
\centering
\includegraphics[width=\textwidth]{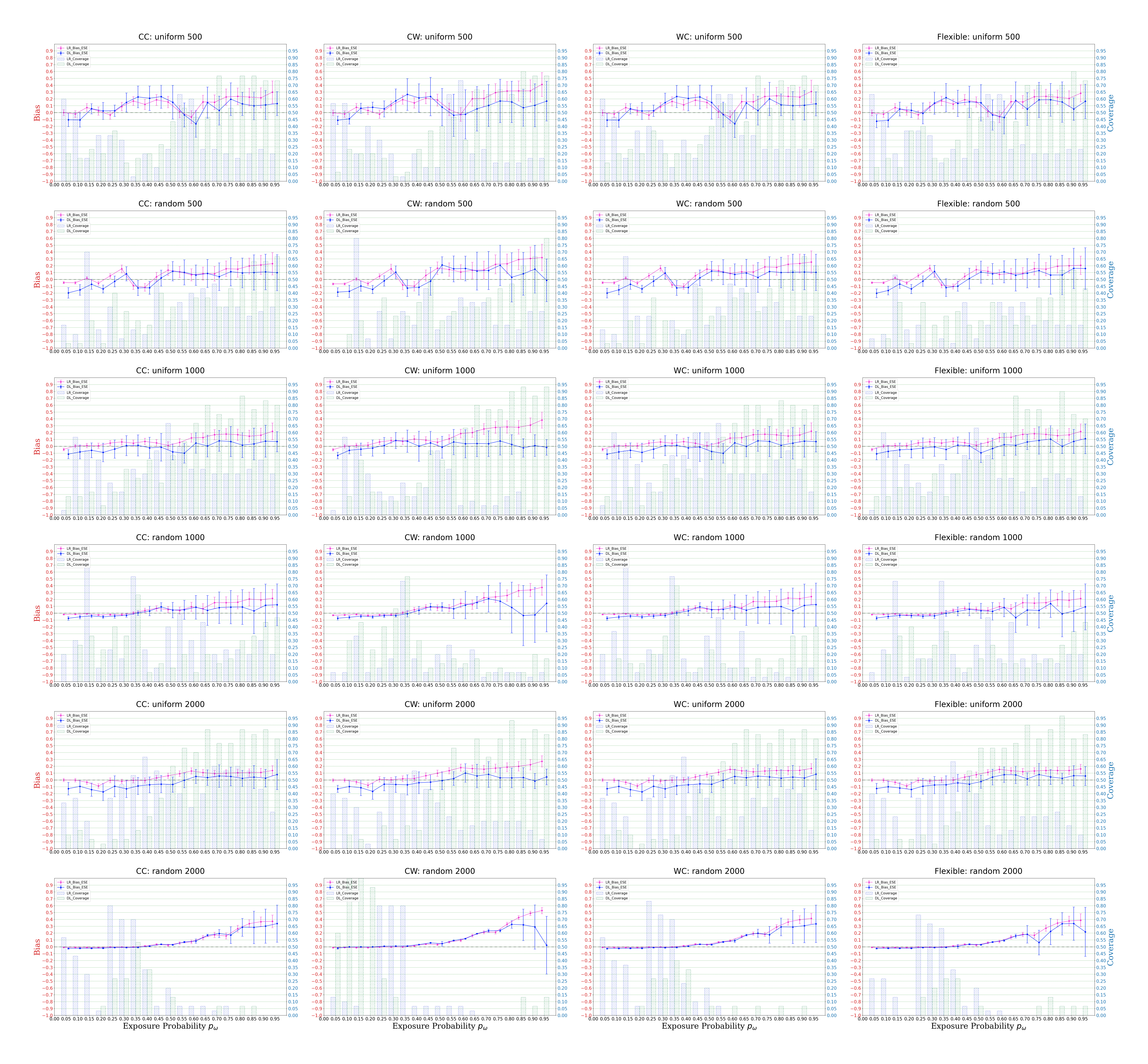}
\caption{Performance comparison between linear regression and deep learning model in predicting the cumulative infliction of disease for uniform and power-law random graph with $N=500, 1000$ and $2000$ under constant strategy with budget covering $50\%$ population and most connected are exposed. The deep learning model has a reception field over the last ten time steps of observational data. Bias and ESE are viewed using the left y-axis in red. The cover is shown in blue on the right y-axis. The grey horizontal dash indicated the zero as a benchmark line for bias estimation.}
\label{fig:results_node_top50}
\end{figure*}

As $N$ increases for both uniform and power-law random graphs, we observe a consistent trend where bias increases with higher exposure probabilities. However, the ESE for both linear regression and deep learning models tends to decrease as $N$ increases. In terms of coverage, linear regression models show a decreasing trend as $N$ increases for power-law random graphs, while the coverage from deep learning models remains relatively stable across different network sizes. Deep learners are better able to capture the true mean of the outcome within the 95\% confidence interval for exposure probabilities greater than 0.3, across all values of $N$ ($500, 1000, 2000$). This suggests that deep learning models are more robust in predicting counterfactuals, especially when they are further shifted from the observed data, providing more accurate and stable predictions in larger networks.

\subsubsection{Strategy: constant with budget covering $50\%$ individuals}
\noindent Figure \ref{fig:results_node_top50} illustrates the performance of outcome prediction under constant strategy, assuming there is enough budget covering $50\%$ individuals, and we choose to quarantine the most connected individuals.


Similar to the scenario where the entire population is covered, we observe that deep learners consistently achieve smaller bias and ESE compared to linear regression models. However, the threshold for noticeable bias improvement in deep learners shifts to a lower exposure probability of $p_{\omega} < 0.5$. This indicates that, when quarantine measures affect only half of the population, the distribution shift is less pronounced, allowing the linear model to adequately capture the effects at lower exposure probabilities. As the exposure probability increases, however, the deep learner’s ability to account for the more significant distribution shift between the sampled data and the observed data becomes crucial for maintaining prediction accuracy.

For exposure probabilities greater than $0.5$, the deep learner’s coverage consistently outperforms that of the linear regression model. This is particularly evident when comparing the performance in power-law random graphs, as shown in Figure \ref{fig:results_node_top50}. In the scenario where only half of the population is quarantined, we observe that misspecification in the outcome model leads to performance deterioration in both linear regression and deep learning models. Although the deep learner provides better coverage than linear regression, for exposure probabilities greater than $0.5$, the coverage drops below $0.4$, which suggests that neither model remains a robust estimator under these conditions. This trend is also reflected in the latent coverage comparison.

Different from the strategy that has a budget covering all populations, the performance improvement of the deep learners is more significant in power-law random graphs than in uniform graphs at a higher exposure probability, which can be attributed to the fact that the application of quarantine on 50\% of the population will not change the data distribution significantly under lower exposure probability. Misspecification in the exposure or outcome model variables has a minimal effect on deep learners but significantly degrades the performance of linear regression models under $p_{\omega} > 0.5$, particularly in the CW and WC scenarios. Additionally, as network size increases, the models’ bias increases, but deep learners exhibit improved ESE and consistent coverage, especially in power-law graphs. Larger networks allow deep learners to better generalize and handle distribution shifts, outperforming linear regression models as the exposure probability rises.


Supplementary Figure 2 illustrates the performance of outcome prediction under constant strategy, assuming there is enough budget covering $50\%$ individuals, and we choose to quarantine the least connected individuals.

When quarantine is applied to the least connected individuals, the distribution drift between the counterfactual and the factual data is significantly reduced compared to strategies targeting the most connected individuals. Consequently, the improvement in bias estimation for DeepNetTMLE is not as pronounced. However, the trend of deep learners demonstrating more robust performance at higher exposure levels ($p_{\omega} > 0.7$)  persists. Notably, under this strategy, benchmark linear regression and L2 models are more prone to overestimating the counterfactual, leading to positive bias, while DeepNetTMLE tends to underestimate the counterfactual, resulting in negative bias. This is a unique pattern compared to the ones observed in strategies targeting the most connected nodes or the entire population. The reduced distribution drift makes it harder for the domain adaptation training to distinguish between the target and source domains, weakening its effectiveness and contributing to bias underestimation. Moreover, the bias observed under this strategy is the smallest across all tested strategies for both benchmark models and DeepNetTMLE. This suggests that targeting the least connected individuals introduces less variability into counterfactual predictions, yielding more consistent and stable estimates across models, albeit with less room for improvement by advanced methods like DeepNetTMLE.

For high exposure probabilities ($p_{\omega} > 0.5$), DeepNetTMLE consistently produces better coverage and latent coverage compared to benchmark models. However, at smaller exposure probabilities, benchmark models demonstrate superior coverage, with DeepNetTMLE even reaching zero coverage in extreme cases. This is likely due to the limited treatment probability at $p_{\omega} < 0.5$, which minimizes the difference between sampled and observed data, allowing linear regression models to provide sufficiently accurate counterfactual estimations. For latent coverage, DeepNetTMLE demonstrates robust performance by effectively controlling the dependencies between observations to the second-hop neighborhood. This robustness highlights the advantage of DeepNetTMLE in capturing network interactions and spillover effects under temporal confounding that are beyond the capability of traditional regression models.

The DeepNetTMLE demonstrates consistent performance for uniform graphs but shows limitations for power-law random graphs, particularly with random graphs of $N=500$, where deep learners fail to outperform benchmark models across all metrics. This suggests that graph connectivity can significantly impact model performance. However, under misspecification of input variables, DeepNetTMLE provides robust estimates, consistently achieving smaller bias and larger coverage for all graphs when $p_{\omega} > 0.5$. The number of nodes in the graphs does not negatively impact model performance. Instead, improvements in bias and coverage over benchmarks are consistently observed as the graph size increases. Notably, for random graphs with $N=2000$, both DeepNetTMLE and benchmark models achieve minimal bias. This phenomenon may occur because, with larger adjacency matrices, interventions targeting the least connected nodes minimally influence epidemic progression, reducing the difference between counterfactual and factual outcomes and simplifying the estimation process.

\subsection{Ablation Studies}

\begin{figure}[!h]
\centering
\includegraphics[width=\linewidth]{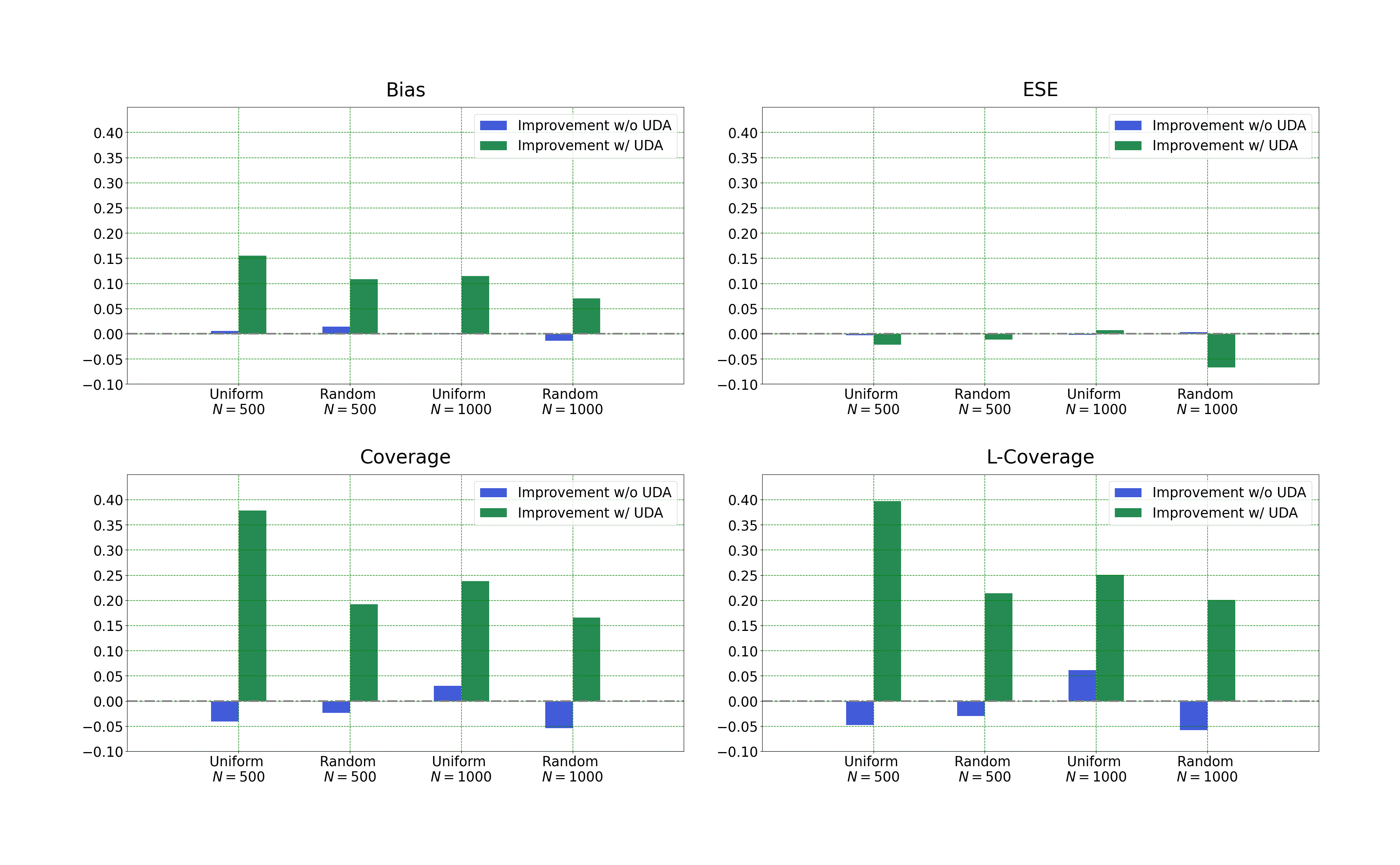}
\caption{Average performance improvement comparing DeepNetTMLE over linear regression w/ and w/o unsupervised domain adaptation, respectively. The results are averaged over four scenarios (CC, CW, WC, Flexible) and all exposure probability levels. The grey horizontal line indicated the zero as a benchmark line for improvement.}
\label{fig:abl_uda}
\end{figure}

\noindent The domain adaptation module is pivotal in improving model performance by addressing time-varying dependencies. In Figure \ref{fig:abl_uda}, we compare the performance of the model trained with and without the domain adaptation module across four metrics and four scenarios for uniform and power-law random graphs with $N=500, 1000$. For bias improvement, models incorporating domain adaptation show more significant gains over the benchmark linear regression models, though models without domain adaptation also achieve notable improvements. However, the model without domain adaptation demonstrates less variability in its estimations. This variability stems from the competing objectives in the domain adaptation loss function: enhancing source domain classification accuracy while reducing domain classification accuracy, which can introduce fluctuations in target domain performance. In terms of coverage and latent coverage, the model with domain adaptation consistently outperforms its counterpart, delivering substantially better results. This underscores the importance of removing time-varying dependencies in the deep learning outcome model within the TMLE framework. Such removal is essential for generating reliable counterfactual predictions under general interference scenarios. These findings emphasize that domain adaptation is a critical feature for robust and accurate deep learning-based causal inference.

\begin{figure}[!h]
\centering
\includegraphics[width=\linewidth]{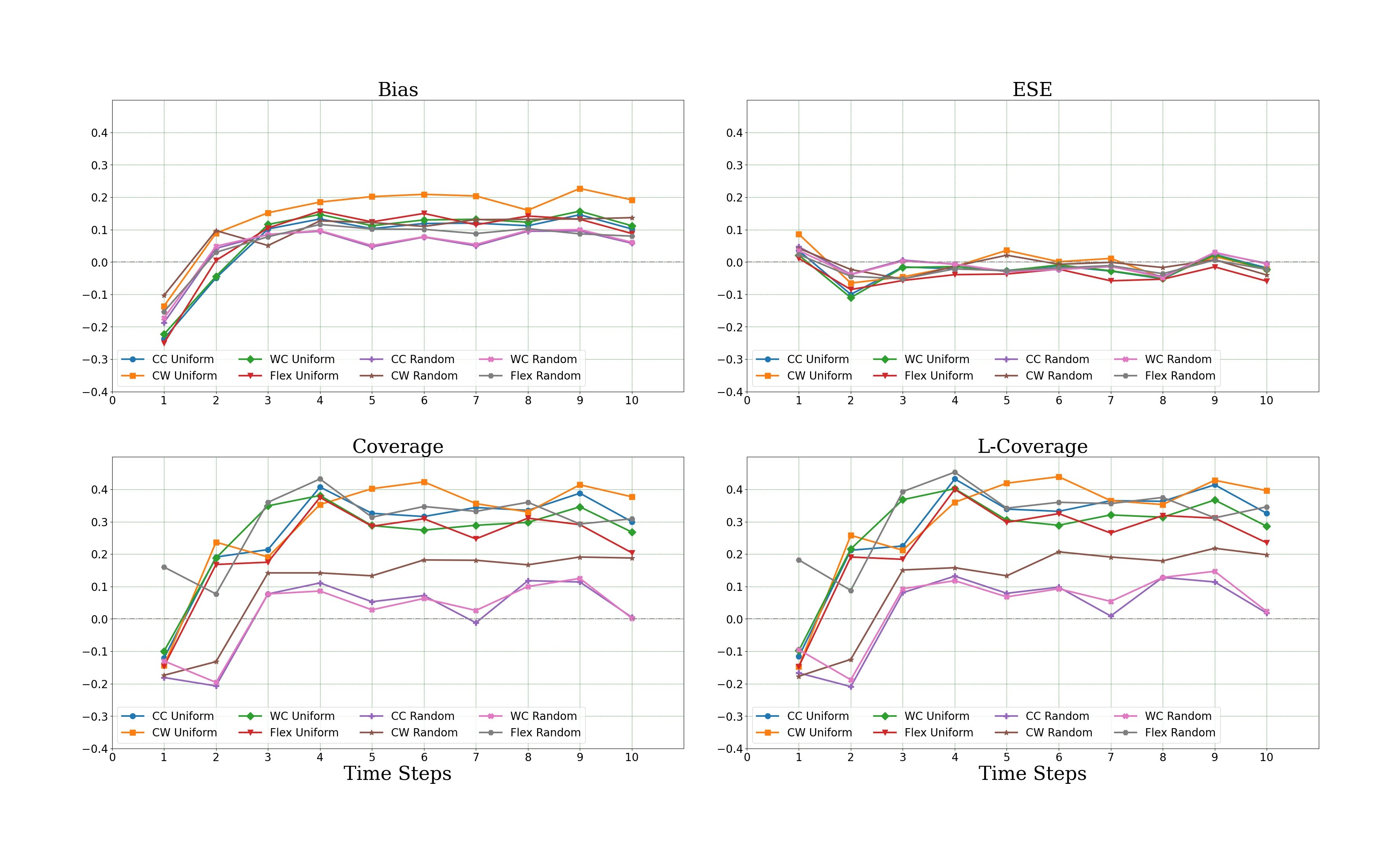}
\caption{Metric Improvement of DeepNetTMLE over Linear Regression for different reception fields, including the last 1 to 10 steps.  The Experiments are run using uniform and power-law random graphs with $N=500$ under CC, CW, WC, and flexible scenarios.}
\label{fig:abl_ts}
\end{figure}

We evaluate the performance of deep learning models relative to linear regression models using varying lengths of observed data, as detailed in Figure \ref{fig:abl_ts}. When using only the last 1 or 2 time steps, deep learning models tend to underperform due to underfitting, resulting in worse bias estimation compared to linear regression. However, with a reception field of $T \geq 3$, deep learning models consistently outperform linear regression in bias estimation, coverage, and latent coverage, even when model input variables are misspecified. DeepNetTMLE performs best for $T=9$, achieving significant improvements across the bias, ESE, coverage, and latent coverage. This highlights the importance of a reception field that sufficiently captures the epidemic progression. However, extending the field to $T=10$ deteriorates performance, likely because the sampled data for $T=1$ (common to both observed and sampled datasets) introduces misleading signals that disrupt accurate counterfactual predictions.

In uniform graphs, the bias improvement of DeepNetTMLE over benchmark models becomes more significant when the reception field is $T \geq 3$. Additionally, coverage and latent coverage improvements are more pronounced in uniform graphs under the CC, CW, and WC scenarios. However, in the flexible scenario, DeepNetTMLE demonstrates better coverage and latent coverage in power-law random graphs. Overall, the performance trends of the model are consistent across both uniform and power-law random graphs, with slight variations depending on the scenario. DeepNetTMLE’s performance is particularly notable under outcome model input variable misspecification and flexible scenarios, implicitly handling feature selection effectively when used as the outcome model. Despite these strengths, deep learners exhibit a slight disadvantage in ESE compared to linear regression models, though the difference is minimal. This underscores that while deep learning models excel in bias and coverage accuracy, linear models may still yield slightly more consistent outcome variations.

\begin{figure}[!h]
\centering
\includegraphics[width=\linewidth]{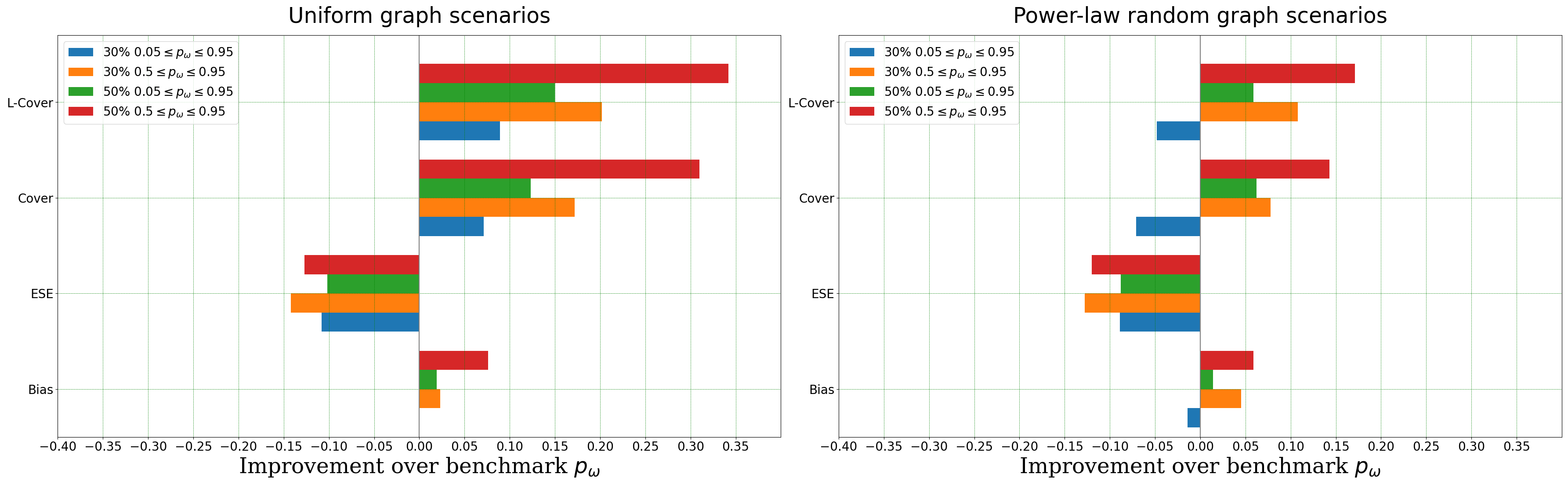}
\caption{Average performance improvement comparing DeepNetTMLE over linear regression under 30\% and 50\% budget constraint, respectively. The results are averaged over four scenarios (CC, CW, WC, Flexible) and exposure probability levels. The grey vertical line indicated the zero as a benchmark line for improvement.}
\label{fig:abl_30_vs_50}
\end{figure}

To validate the results under different budget constraints, we conduct experiments on datasets where budgets covering 30\% and 50\% of the population are simulated, with the most connected nodes selected for quarantine. The objective is to demonstrate that varying budget assumptions do not significantly affect the improvement of DeepNetTMLE over benchmark models at higher exposure probabilities. The results are shown in Figure \ref{fig:abl_30_vs_50}. For $p_{\omega} \geq 0.5$, DeepNetTMLE consistently outperforms the benchmarks in bias reduction, achieving average improvements of 0.034 and 0.067 for 30\% and 50\% budget scenarios, respectively, across all four scenarios for uniform and random graphs with N=500. For a 50\% budget, the improvement persists across all exposure probabilities (0.016), but for a 30\% budget, there is a slight negative impact (-0.007). In terms of coverage, the 50\% budget shows better performance across all exposure probabilities, with a net increase of 0.093, and a more pronounced improvement of 0.226 for $p_{\omega} \geq 0.5$. The 30\% budget achieves no overall improvement for all exposure probabilities but does see an increase of 0.125 for $p_{\omega} \geq 0.5$. Latent coverage improves consistently under both budget plans, with average increases of 0.021 for 30\% and 0.105 for 50\% coverage across all exposure probabilities, and even larger improvements of 0.155 and 0.256, respectively, for $p_{\omega} \geq 0.5$. Regarding ESE, linear regression benchmark models demonstrate more consistent results than DeepNetTMLE, with deep learner variation increasing as exposure probability rises. Based on these findings, we recommend that for smaller interventions affecting a limited population ($p_{\omega} < 0.5$) under a 30\% budget, benchmark models are adequate for estimating the ATE. However, for larger interventions ($p_{\omega} \geq 0.5$) with 50\% policy coverage, DeepNetTMLE provides superior counterfactual predictions.


\section{Conclusion}
\noindent This study addressed the critical challenge of causal inference under interference, a core issue in infectious disease control. Unlike traditional approaches that assume no interference between individuals (SUTVA), our work leverages a longitudinal framework to capture time-varying confounding and interference effects in the context of disease transmission. We proposed a novel model, DeepNetTMLE, which combines domain adaptation to account for time dependence and targeting techniques to achieve stable, accurate counterfactual predictions. By simulating a realistic “leaky” quarantine scenario through the SIR model, we quantified the spillover effects on disease outcomes and evaluated model performance in scenarios with full and partial quarantine budgets.

Our experiments demonstrated that DeepNetTMLE outperformed traditional approaches like linear and L2-penalized regression models, yielding lower bias, smaller ESE, and higher coverage, even under model misspecifications. This robustness makes DeepNetTMLE particularly promising for estimating causal effects in public health contexts where randomized trials are infeasible. The model’s counterfactual predictions offer valuable insights for optimizing quarantine strategies, such as determining ideal quarantine probabilities and targeting the most connected individuals cost-effectively.

Future work will extend DeepNetTMLE’s applicability by (1) validating its predictions on real-world datasets and (2) establishing theoretical guarantees for its effectiveness in causal inference with longitudinal data and interference. Through these advances, DeepNetTMLE holds the potential to become a vital tool for guiding data-driven public health interventions.

\section*{Acknowledgments}
\noindent This work was supported in part by the STI 2030-Major Projects of China under Grant 2021ZD0201300, the National Science Foundation of China under Grant 62276127, and the Fundamental Research Funds for the Central Universities under Grant 2024300394.

\bibliographystyle{IEEEtran}
\bibliography{DeepNetTMLE}



\end{document}